\documentclass[10pt,twocolumn,letterpaper]{article}

\usepackage{cvpr}              % To produce the CAMERA-READY version
\usepackage[dvipsnames]{xcolor}

\definecolor{cvprblue}{rgb}{0.21,0.49,0.74}
\usepackage[pagebackref,breaklinks,colorlinks,citecolor=cvprblue]{hyperref}
\usepackage{xspace}
\usepackage{multirow}
\usepackage{makecell}
\usepackage{bbm}
\usepackage{bm}
\usepackage[T1]{fontenc}
\newtheorem{proposition}{Proposition}

\usepackage{algorithm}
\usepackage{algorithmic}
\usepackage[accsupp]{axessibility} % Improves PDF readability for those with visual impairments.

\newcommand{\smallsection}[1]{{\vspace{0.05in} \noindent \bf {#1.\hspace{5pt}}}}
\newcommand{\avgenc}{Avg.Encoder\xspace}
\newcommand{\aum}{AUM\xspace}
\newcommand{\cl}{CL\xspace}
\newcommand{\cores}{CORES\xspace}
\newcommand{\simifeat}{SIMIFEAT\xspace}
\newcommand{\proposed}{DynaCor\xspace}

\newcommand{\divmix}{Dividemix\xspace}
\newcommand{\divl}{DDyna-L\xspace}
\newcommand{\divs}{DDyna-S\xspace}
\def\paperID{11908} % *** Enter the Paper ID here
\def\confName{CVPR}
\def\confYear{2024}

\title{Learning Discriminative Dynamics with Label Corruption for Noisy Label Detection}

\author{
Suyeon Kim$^{1}$, Dongha Lee$^{2}$\thanks{Corresponding authors}, SeongKu Kang$^{3}$, Sukang Chae$^{1}$, Sanghwan Jang$^{1}$, Hwanjo Yu$^{1}$\footnotemark[1]\\ 
$^{1}$ POSTECH, 
$^{2}$ Yonsei University,  
$^{3}$ University of Illinois at Urbana Champaign\\
{\tt\small \{kimsu, chaesgng2, s.jang, hwanjoyu\}@postech.ac.kr, donalee@yonsei.ac.kr, seongku@illinois.edu
}
}

\begin{document}
\maketitle

\begin{abstract}
Label noise, commonly found in real-world datasets, has a detrimental impact on a model's generalization.
To effectively detect incorrectly labeled instances, previous works have mostly relied on distinguishable training signals, such as training loss, as indicators to differentiate between clean and noisy labels.
However, they have limitations in that the training signals incompletely reveal the model's behavior and are not effectively generalized to various noise types, resulting in limited detection accuracy.
In this paper, we propose \proposed framework that distinguishes incorrectly labeled instances from correctly labeled ones based on the dynamics of the training signals.
To cope with the absence of supervision for clean and noisy labels, \proposed first introduces a label corruption strategy that augments the original dataset with intentionally corrupted labels, enabling indirect simulation of the model's behavior on noisy labels.
Then, \proposed learns to identify clean and noisy instances by inducing two clearly distinguishable clusters from the latent representations of training dynamics. 
Our comprehensive experiments show that \proposed outperforms the state-of-the-art competitors and shows strong robustness to various noise types and noise rates.
\end{abstract}

\section{Introduction}
\label{sec:intro}

The remarkable success of deep neural networks (DNNs) is largely attributed to massive and accurately labeled datasets.
However, creating such datasets is not only expensive but also time-consuming. 
As a cost-effective alternative, various methods have been employed for label collection, such as crowdsourcing \cite{deng2009imagenet} and extracting image labels from accompanying text on the web \cite{xiao2015learning, li2017webvision}.
Unfortunately, these approaches have led to the emergence of noise in real-world datasets, with reported noise rates ranging from 8.0\% to 38.5\% \cite{xiao2015learning, li2017webvision, lee2018cleannet}, which severely degrades the model's performance \cite{zhang2021understanding, arpit2017closer}.

To cope with the detrimental effect of such noisy labels, a variety of approaches have been proposed, including noise robust learning that minimizes the impact of inaccurate information from noisy labels during the training process~\cite{liu2020early, wei2021open, xiao2015learning, cheng2020learning} and data re-annotation through algorithmic methods \cite{song2019selfie, han2019deep, zhang2020distilling}.
Among them, the task of noisy label detection, which our work mainly focuses on, aims to identify incorrectly labeled instances in a training dataset~\cite{cheng2020learning, pleiss2020identifying, kim2021fine}.
This task has gained much attention in that it can be further utilized for improving the quality of the original dataset via cleansing or rectifying such instances.

Motivated by the \textit{memorization effect}, which refers to the phenomenon where DNNs initially grasp simple and generalized patterns in correctly labeled data and then gradually overfit to incorrectly labeled data~\cite{arpit2017closer}, most existing studies have utilized distinguishable training signals as indicators of label quality to differentiate between clean and noisy labels.
To elaborate, these training signals are derived from the model's behavior on individual instances during the training~\cite{swayamdipta2020dataset, wang2022deep}, involving factors such as training loss or confidence scores. 
Note that it is impractical to acquire annotations explicitly indicating whether each instance is correctly labeled or not.
Hence, numerous studies have crafted various heuristic training signals~\cite{forouzesh2023differences, huang2019o2u, kim2021fine}, designed based on human prior knowledge of the model's distinctive behaviors when faced with clean and noisy labels.

Despite their effectiveness, the training signal-based detection methods still exhibit several limitations:
(1) They only focus on a scalar signal at a single epoch (or a representative one across the entire training trajectory), which leads to limited detection accuracy (See Appendix B.2).
Since the model's distinct behaviors on clean and noisy labels draw different temporal trajectories of training signals, a single scalar is insufficient to distinguish them by capturing temporal patterns within training dynamics.
(2) Existing detection approaches based on heuristics are not effectively generalized to various types of label noise.
Noisy labels can originate from diverse sources, including human annotator errors \cite{peterson2019human, wei2021learning}, systematic biases~\cite{wang2021policy}, and unreliable annotations from web crawling~\cite{xiao2015learning}, resulting in different noise types and rates for each dataset;
this eventually requires considerable efforts to tune hyperparameters for training recipes of DNNs~\cite{li2020dividemix, liu2020early, wang2022promix}.

To tackle these challenges, our goal is to propose a fully data-driven approach that directly learns to distinguish the training dynamics of noisy labels from those of clean labels using a given dataset without solely relying on heuristics.
The primary technical challenge in this data-driven approach arises from the absence of supervision for clean and noisy labels.
As a solution, we introduce a \textit{label corruption} strategy--image augmentation attaching intentionally corrupted labels via random label replacement.
Since the augmented instances are highly likely to have incorrect labels, we can utilize them to capture the training dynamics of noisy labels.
In other words, this allows us to simulate the model's behavior on noisy labels by leveraging the augmented instances with corrupted labels.

In this work, we present a novel framework, named \proposed, that learns discriminative \textbf{Dyna}mics with label \textbf{Cor}ruption for noisy label detection.
To be specific, \proposed identifies clean and noisy labels via clustering of latent representations of training dynamics. 
To this end, it first generates training dynamics of original instances and corrupted instances.
Then, it computes the dynamics representations that encode discriminative patterns within the training trajectories by using a parametric dynamics encoder.
The dynamics encoder is optimized to induce two clearly distinguishable clusters (i.e., each for clean and noisy instances) based on two different types of losses for
(1) high cluster cohesion and (2) cluster alignment between original and corrupted instances.
Furthermore, \proposed adopts a simple validation metric for the dynamics encoder based on the clustering quality so as to indirectly estimate its detection performance where ground-truth annotations of clean and noisy labels are not available for validation as well.

The contribution of this work is threefold as follows:
\begin{itemize}
    \item We introduce a label corruption strategy that augments the original data with corrupted labels, which are highly likely to be noisy, enabling indirect simulation of the model's behavior on noisy labels during the training.
    
    \item We present a data-driven \proposed framework to distinguish incorrectly labeled instances from correctly labeled ones via clustering of the training dynamics.

    \item Our extensive experiments on real-world datasets demonstrate that \proposed achieves the highest accuracy in detecting incorrectly labeled instances and remarkable robustness to various noise types and noise rates.    
\end{itemize}

\section{Related Work}
\label{sec:related}

\begin{figure*}[t]
    \centering
    \includegraphics[width=\linewidth]{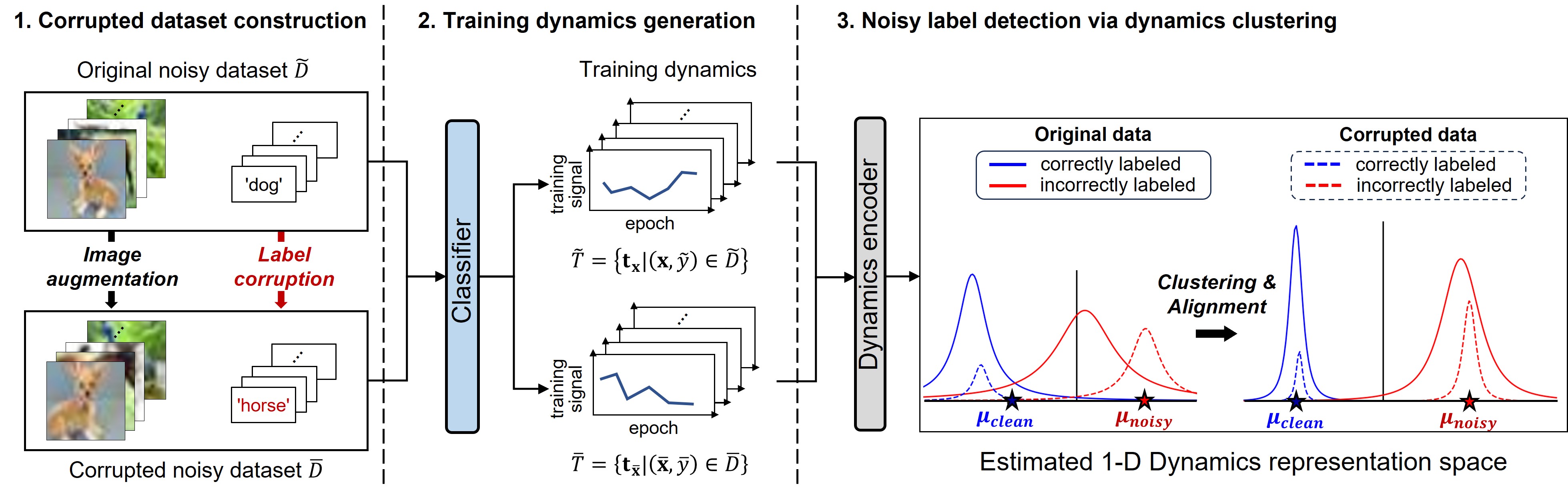}
    \caption{The proposed \proposed framework consists of three steps: (1) Corrupted dataset construction generates the augmented images with corrupted labels, likely resulting in noisy labels, in order to provide guidance for discrimination between clean and noisy labels.  
    (2) Training dynamics generation collects the trajectory of training signals for both the original and corrupted datasets by training a classifier. (3) Noisy label detection is performed by discovering two distinguishable clusters of dynamics representations, and for this, the dynamics encoder is optimized to enhance both cluster cohesion and alignment between the original and the corrupted datasets.}
    \label{fig:arch}
\end{figure*}

We provide a brief overview of the two primary research directions for addressing incorrectly labeled instances in a noisy dataset:
(1) \textit{Noisy label detection} focuses on identifying instances that are incorrectly labeled within a dataset, aiming to enhance data quality.
(2) \textit{Noise robust learning} is centered on developing learning algorithms and models that are resilient to the impact of noisy labels, ensuring robust performance even in the presence of labeling errors.

\smallsection{Noisy label detection}
The main challenge in detecting noisy labels lies in defining a surrogate metric for label quality, essentially indicating how likely an instance is correctly labeled.
The widely adopted option is the training loss, assessing the disparity between the model prediction and given labels~\cite{jiang2018mentornet, han2018co, huang2019o2u}, with higher loss often indicating incorrect labels. 
Various proxy measures, including gradient-based values~\cite{zhang2018generalized, wang2019symmetric} and prediction-based metrics~\cite{northcutt2021confident, sun2020crssc, song2019selfie, pleiss2020identifying} have been developed to differentiate between clean and noisy labels, utilizing methods like Gaussian mixture models~\cite{zoran2011learning, li2020dividemix, kim2021fine, chen2023sample} or manually designed thresholds~\cite{northcutt2021confident,han2018co, yu2019does, zhu2022detecting, pleiss2020identifying}. 
However, these approaches may overlook the potential benefits of adopting a data-driven (or learning-centric) detection model ~\cite{cheng2020learning}, which can be easily generalized to various noise types and levels. 
As a training-free alternative, a recent study~\cite{zhu2022detecting} introduces a non-parametric KNN-based approach based on the assumption that instances situated closely in the input feature spaces derived from a pre-trained model are more likely to share the same clean label. 
However, its efficacy in detection heavily depends on the quality of the pre-trained model and may not be universally applicable across domains with specific fine-grained visual features.

\smallsection{Noise robust learning}
Extensive research have focused on creating noise robust methods: loss functions \cite{zhang2018generalized, wang2019symmetric}, regularization~\cite{liu2020early, cheng2021mitigating, cheng2022class}, model architectures~\cite{xiao2015learning, chen2015webly, bekker2016training, jindal2016learning, goldberger2016training, yao2018deep, cheng2020weakly}, and training strategies~\cite{zhang2017mixup, xia2020robust, lukasik2020does, kim2021comparing}.
Recent studies have endeavored to integrate the process of detecting noisy labels and appropriately addressing them into the training pipeline in various ways: re-weighting losses~\cite{jiang2018mentornet, ren2018learning, shu2019meta} or re-annotation~\cite{song2019selfie, han2019deep, zhang2020distilling}. 
Besides, several studies~\cite{li2020dividemix, wang2022promix, wei2022self, chen2023sample} treat detected noisy labels as unlabeled and make use of established semi-supervised techniques ~\cite{han2019deep, zhang2020distilling, zhang2017mixup, berthelot2019mixmatch}.
Current robust learning typically relies on clean data, i.e., test data, for validation, while noisy detection methods can function without it, making direct comparisons difficult~\cite{zhu2022detecting}. 
In this sense, we will discuss how these noise robust learning approaches can be effectively combined with noisy detection methods (Sec.~\ref{subsec:companal}).

\section{Problem Formulation}
\label{sec:pre}
% \smallsection{Noisy Label Detection}
For multi-class classification, let $\mathcal{X}$ be an input feature space and $\mathcal{Y}=\{1,2,.., C\}$ be a label space.
Consider a dataset $D=\{(\mathbf{x}_n, y_n)\}^N_{n=1}$, where each sample is independently drawn from an unknown joint distribution over $\mathcal{X}\times\mathcal{Y}$.
In real-world scenarios, we can only access a noisily labeled training set $\widetilde{D}=\{(\mathbf{x}_n, \tilde{y}_n)\}^N_{n=1}$, where $\tilde{y}$ denotes a noisy annotation, and there may exist $n\in\{1,..., N\}$ such that $y_n\neq\tilde{y}_n$.  
In this work, we focus on the task of \textit{noisy label detection}, which aims to identify the incorrectly labeled instances, i.e., $\{(\mathbf{x}_n, \tilde{y}_n)\in\widetilde{D} \mid y_n \neq \tilde{y}_n\}$.
As an evaluation metric, we use F1 score \cite{lipton2014optimal}, treating the incorrectly labeled instances as positive and the remainings as negative. 

\section{Methodology}
\label{sec:method}

\subsection{Overview}
\proposed (\textbf{Dyna}mics learning with label \textbf{Cor}ruption for noisy label detection) framework learns discriminative patterns inherent in training dynamics, thereby distinguishing incorrectly labeled instances from clean ones. 
As illustrated in Figure \ref{fig:arch}, \proposed consists of three major steps.
\begin{itemize}
    \item Corrupted dataset construction (Sec.~\ref{sec:corruption}):
    To address the challenge arising from the lack of supervision for incorrectly labeled instances, we introduce a corrupted dataset that intentionally corrupts labels, providing guidance to identify incorrectly labeled instances. 
    
    \item Training dynamics generation (Sec.~\ref{sec:dynamics}):
    We generate training dynamics, which denote a model's behavior on individual instances during training, by training a classifier using both the original and the corrupted dataset.
    
    \item  Noisy label detection via dynamics clustering (Sec.~\ref{sec:detection}):
    We seek to discover underlying patterns in the training dynamics by learning representations that reflect the intrinsic similarities among data points, leveraging the characteristics of the corrupted dataset.	
    For this, we encode the training dynamics via a dynamics encoder that learns discriminative representation using clustering and alignment losses.   
    Then we find clusters using a robust validation metric designed for dynamics-based clustering.
\end{itemize}
\subsection{Corrupted dataset construction}
\label{sec:corruption}

Given the original dataset $\widetilde{D}$, we construct a corrupted dataset $\bar{D}$ by intentionally corrupting labels for a randomly sampled subset of $\widetilde{D}$ with a corruption rate $\gamma \in (0,1]$. 
Specifically, to obtain a corrupted instance $(\bar{\mathbf{x}}, \bar{y})$ from an original data instance $(\mathbf{x}, \tilde{y})$, we transform an input image using weak augmentation such as horizontal flip or center crop, i.e., $\bar{\mathbf{x}}=\mathrm{Aug}(\mathbf{x})$.
Then, we randomly flip the class label to one of the other classes, i.e., $\bar{y}\in\{1,..., C\}\backslash\{\tilde{y}\}$.
The corrupted dataset, guaranteed to exhibit symmetric noise at a higher rate than the original, provides additional signals for discerning incorrectly labeled instances in the clustering process, as detailed in the following analysis.

% It is guaranteed that the corrupted dataset always has a symmetric type noise with a higher noise rate than the original dataset, as we will explain in the following analysis.
% Based on this property, we leverage the corrupted dataset to yield additional signals to learn distinctive patterns of incorrectly labeled instances in the subsequent clustering.

\smallsection{Analysis: the noise rate of the corrupted dataset}
We analyze the lower bound on the noise rate of the corrupted dataset $\bar{D}$. 
Let $\eta \in[0,1]$ denote the noise rate of the original dataset $\widetilde{D}$.\footnote{$\eta=\frac{1}{\vert\widetilde{D}\vert}{\vert\{(\mathbf{x},\tilde{y})\in\widetilde{D} \mid \tilde{y} \neq y,\ (\mathbf{x},y) \in D\}\vert}$}
Following the previous literature \cite{sukhbaatar2014training,han2018co, gui2021towards}, we presume the \textit{diagonally dominant condition}, i.e., $\mathrm{Pr}(\tilde{y}=i|y=i) > \mathrm{Pr}(\tilde{y}=j|y=i), \forall i\neq j$, which indicates that correct labels should not be overwhelmed by the false ones. 
With this condition of $\eta < 1-\frac{1}{C}$, we have the following proposition.

\begin{proposition}[Lower bound of $\eta_\gamma$]
\label{pro1}
Let $\eta_{\gamma}$ denote the noise rate of the corrupted dataset. 
Given the diagonally dominant condition, i,e., $\eta < 1-\frac{1}{C}$,  for any $\gamma\in{\left(0,1\right]}$, $\eta_{\gamma}$ has a lower bound of $1-\frac{1}{C}$.
\end{proposition}
The proof is presented in Appendix C, from which we can derive $\eta < \eta_{\gamma}$.

\subsection{Training dynamics generation}
\label{sec:dynamics}

\subsubsection{Training dynamics} 
The training dynamics indicates a model's behavior on individual instances during the training, quantitatively describing the training process \cite{swayamdipta2020dataset, wang2022deep}. 
Concretely, the training dynamics is defined as the trajectory of training signals derived from a model's output across the training epochs.  
In the literature, various types of training signals~\cite {zhou2020curriculum, swayamdipta2020dataset, arpit2017closer} have been employed for analyzing the model's behavior.

Given a classifier $f$, let $f(\mathbf{x})\in \mathbb{R}^C$ denote the output logits of an instance $\mathbf{x}$ for $C$ classes.  
Let $t$ be a transformation function that maps $C$ logits to a scalar training signal. 
In this paper, we use \textit{quantized logit difference} as the training signal.\footnote{We provide a detailed analysis of various training signals for identifying incorrectly labeled instances in Appendix B.3}
It quantizes the difference between a logit ~\cite{pleiss2020identifying} of a given label and the largest logit among the remaining classes, i.e., $t(f(\mathbf{x}), \tilde{y})= \text{sign}(f_{\tilde{y}}(\mathbf{x}) - \max_{c\neq \tilde{y}}f_c(\mathbf{x})),$ 
where $f_c(\mathbf{x})$ denotes the logit for class $c$, and $\text{sign}(\mathbf{x})=1$ or -1 if $\mathbf{x}>=0$ or $<0$, respectively. 
The training dynamics for an instance $\mathbf{x}$ is defined as
\begin{equation}
\label{eq:dynamics_trajectories}
     \mathbf{t}_{\mathbf{x}}=[t^{(1)}(f(\mathbf{x}),\tilde{y}),..,t^{(E)}(f(\mathbf{x}),\tilde{y})], 
\end{equation}
where $t^{(e)}(f(\mathbf{x}),\tilde{y})$ denotes the training signal computed at epoch $e$, and $E$ is the maximum number of training epochs.
For the sake of convenience, we denote $\mathbf{t}_{\mathbf{x}}$ and $t^{(e)}_{\mathbf{x}}$ as an abbreviation for $\mathbf{t}(\mathbf{x}, \tilde{y};f)$ and $t^{(e)}(f(\mathbf{x}),\tilde{y})$, respectively.

\subsubsection{Dynamics generation for noisy label detection} 
We generate training dynamics for both the original and the corrupted datasets.
Specifically, we train a classifier by minimizing the classification loss on $\widetilde{D}$ and $\bar{D}$:
\begin{equation}
\frac{1}{|\widetilde{D}|}\sum_{(\mathbf{x}, \tilde{y})\in{\widetilde{D}}}{\ell_{ce}(f(\mathbf{x}), \tilde{y})} + 
\frac{1}{|\bar{D}|}\sum_{(\bar{\mathbf{x}}, \bar{y})\in{\bar{D}}}{\ell_{ce}\left(f(\bar{\mathbf{x}}), \bar{y}\right) } ,
\end{equation}
where $\ell_{ce}$ is the softmax cross-entropy loss.
For each instance $\mathbf{x}$, we obtain a training dynamics $\mathbf{t}_{\mathbf{x}}\in\mathbb{R}^E$ as specified in Eq.~\eqref{eq:dynamics_trajectories} by tracking $t^{(e)}_{\mathbf{x}}$ over the course of training epochs $E$.
Training dynamics of the original and the corrupted datasets are denoted by $\widetilde{T}:=\{\mathbf{t}_{\mathbf{x}}|(\mathbf{x}, \tilde{y})\in {\widetilde{D}}\}$ and $\bar{T}:=\{\mathbf{t}_{\bar{\mathbf{x}}}|(\bar{\mathbf{x}},\bar{y})\in {\bar{D}}\}$, respectively.

\subsection{Noisy label detection via dynamics clustering}
\label{sec:detection}
We use a clustering approach to identify incorrectly labeled instances within the original dataset. 
Using a dynamics encoder, we encode the generated dynamics and progressively find clusters of correctly and incorrectly labeled instances in the representation space.
The dynamics clustering iterates two key processes: (1) identifications of incorrectly labeled instances (Sec.~\ref{subsubsec:identifying}), and (2) learning distinct representations for each cluster (Sec.~\ref{subsubsec:clustering}).
The clustering quality is assessed by a newly introduced validation metric by leveraging the corrupted dataset without a clean validation dataset (Sec.~\ref{subsubsec:validating}).

\vspace{-1mm}
\subsubsection{Identification of incorrectly labeled instances}
\label{subsubsec:identifying}
\smallsection{Cluster initialization}
Given a training dynamics $\mathbf{t}_{\mathbf{x}}$, a dynamics encoder generates its representation, i.e., $\mathbf{z}_{\mathbf{x}} = \mathrm{Enc}(\mathbf{t}_{\mathbf{x}}) \in \mathbb{R}^{d_{\mathbf{z}}}$. 
Let $\widetilde{Z}$ and $\bar{Z}$ denote the set of dynamics representations of the original and the corrupted datasets, respectively. 
We first introduce trainable parameters for centroids of noisy and clean clusters, i.e., $\bm{\mu}_{noisy},\,\bm{\mu}_{clean}\in\mathbb{R}^{d_{\mathbf{z}}}$.
We initialize $\bm{\mu}_{noisy}$ as the average representation of the corrupted instances $\bar{Z}$, while $\bm{\mu}_{clean}$ is initialized as the average representation of the original instances $\widetilde{Z}$.
Note that this initialization is conducted only once at the beginning of the dynamics clustering step.

\smallsection{Noisy label identification}
We determine whether each instance $\mathbf{x}$ has been incorrectly labeled based on its assignment probability to the noisy cluster.
The assignment probability is computed based on the similarity between $\mathbf{z_x}$ and the noisy cluster's centroid $\bm{\mu}_{noisy}$.
We employ a kernel function based on the Student's $t$-distribution \cite{van2008visualizing} with one degree of freedom as follows:
{\small
\begin{align}
\nonumber
\label{eq:q_noisy}
    q_{noisy}(\mathbf{z_x})&={
    \frac{{(1 + d(\mathrm{\mathbf{z_x}},\bm{\mu}_{noisy}))^{-1}}}
    {{(1 + d(\mathrm{\mathbf{z_x}},\bm{\mu}_{noisy}))^{-1}} +
     {(1 + d(\mathrm{\mathbf{z_x}},\bm{\mu}_{clean}))^{-1}}
    }}, 
    \\
    q_{clean}(\mathbf{z_x})&=1-q_{noisy}(\mathbf{z_x}),
%\label{eq:q_clean}
\end{align}}%
where $d(\mathbf{a}, \mathbf{b}) = 1-\frac{\langle \mathbf{a},\mathbf{b} \rangle}{||\mathbf{a}||_2 \cdot ||\mathbf{b}||_2 }$.
Based on the assignment probability, we regard an instance as incorrectly labeled when its probability to the noisy cluster is predominant. 
\begin{equation}
\label{eq:inference}
v(\mathbf{z_x}):=  \mathbbm{1}[q_{noisy}(\mathbf{z_x}) > q_{clean}(\mathbf{z_x})], 
\end{equation}
$v(\mathbf{z_x})=1$ indicates that $\mathbf{x}$ is predicted to have a noisy label.

\subsubsection{Learning discriminative patterns in dynamics}
\label{subsubsec:clustering}
We introduce the strategy of inducing two distinguishable clusters (each for correctly and incorrectly labeled instances) in the dynamics representation space. 
We propose two types of losses for (1) high cluster cohesion and (2) cluster alignment between original and corrupted instances.

\smallsection{Clustering loss} 
We introduce a clustering loss to make the clusters more distinguishable.
We enhance cluster cohesion by adjusting each instance's representation to be closer to a centroid through a self-enhancing target distribution.
The target distribution is constructed by amplifying the predicted assignment probability \cite{xie2016unsupervised} as follows:
\begin{align}
\nonumber
    p_{noisy}(\mathbf{z_x})&={\frac{{q_{noisy}^2(\mathbf{z_x})/s_{noisy}}}
    {q_{noisy}^{2}(\mathbf{z_x})/s_{noisy}+ q_{clean}^{2}(\mathbf{z_x})/s_{clean}}
    }, \\
p_{clean}(\mathbf{z_x})&=1-p_{noisy}(\mathbf{z_x}),
\end{align}
where $s_{noisy}=\sum_{\mathbf{z}\in{\widetilde{Z}\cup\bar{Z}}}q_{noisy}(\mathbf{z})$ and $s_{clean}=\sum_{\mathbf{z}\in{\widetilde{Z}\cup\bar{Z}}}q_{clean}(\mathbf{z})$.  
Then, we minimize the KL divergence between the cluster assignment distribution $\mathbf{q}(\mathbf{z_x})=[q_{noisy}(\mathbf{z_x}),\, q_{clean}(\mathbf{z_x})]$ and the target distribution $\mathbf{p}(\mathbf{z_x})=[p_{noisy}(\mathbf{z_x}), \,p_{clean}(\mathbf{z_x})]$ as follows:
\begin{equation}
\label{eq:cluster_loss}
    \mathcal{L}_{cluster} = \sum_{\mathbf{z_x}\in{\widetilde{Z}\cup\bar{Z}}}\mathrm{KL}(\mathbf{p}(\mathbf{z_x})||\mathbf{q}(\mathbf{z_x})).
\end{equation}

\smallsection{Alignment loss} 
We introduce an alignment loss that aligns the representation from each cluster's original and corrupted datasets. 
We hypothesize\footnote{It is theoretically proved in \cite{oyen2022robustness}} that symmetric noise is relatively easy to identify among various noise types with diverse difficulty levels.
Consequently, incorrectly labeled instances in the corrupted dataset exhibit more distinctive dynamics patterns than those in the original data, i.e., a red dashed line is farther away from blue lines than a red line in the 3rd step of Fig.\ref{fig:arch} (left).   
From this perspective, the mismatched noise types between the original and the corrupted datasets positively impact the clustering process by adopting alignment loss, which forces a red line to be aligned with a red dashed line in the 3rd step of Fig.\ref{fig:arch} (right).

Instances in the original dataset predicted as noisy and clean are denoted by $\widetilde{Z}_{noisy}=\{\mathbf{z_x}\in \widetilde{Z} |v(\mathbf{z_x}) =1 \}$ and 
$\widetilde{Z}_{clean}=\{\mathbf{z_x}\in \widetilde{Z}|v(\mathbf{z_x}) =0 \}$, respectively. 
Analogously, for the corrupted dataset, we obtain
$\bar{Z}_{noisy}=\{\mathbf{z_x}\in \bar{Z}|v(\mathbf{z_x}) =1 \}$ and
$\bar{Z}_{clean}=\{\mathbf{z_x}\in \bar{Z}|v(\mathbf{z_x}) =0\}$.  
Then, we employ the alignment loss to reduce the discrepancy between the representations of the original dataset and the corrupted dataset as follows:
{
\begin{align}
\nonumber
    \mathcal{L}_{align}^{n} &= d\Big(
    \frac{1}{|\widetilde{Z}_{noisy}|} \sum_{\mathbf{z_x}\in\widetilde{Z}_{noisy}}\mathbf{z_x},
    \frac{1}{|\bar{Z}_{noisy}|} 
    \sum_{\mathbf{z_x}\in\bar{Z}_{noisy}}\mathbf{z_x}
    \Big), \\
\nonumber    
    \mathcal{L}_{align}^c &= d\Big(
    \frac{1}{|\widetilde{Z}_{clean}|} \sum_{\mathbf{z_x}\in\widetilde{Z}_{clean}}\mathbf{z_x},
    \frac{1}{|\bar{Z}_{clean}|} 
    \sum_{\mathbf{z_x}\in\bar{Z}_{clean}}\mathbf{z_x}
    \Big), \\
\label{eq:cosine_alsign}
    \mathcal{L}_{align} &= {\frac{1}{2}}(\mathcal{L}_{align}^n + \mathcal{L}_{align}^c). 
\end{align}
}

\smallsection{Optimization} 
To sum up, the dynamics encoder is optimized by minimizing the following loss:
\begin{equation}
\label{eq:loss}
    \mathcal{L}= \mathcal{L}_{cluster} +\alpha \mathcal{L}_{align},
\end{equation}
where $\alpha$ is a hyperparameter that controls the impact of the alignment loss.

\subsubsection{Validation metric}
\label{subsubsec:validating}
One practical challenge in training the dynamics encoder is determining an appropriate stopping point in the absence of ground-truth annotations of clean and noisy labels for validation.
As a solution, we introduce a new validation metric for the dynamics encoder to estimate its detection performance indirectly.  % based on the clustering quality.
For noisy label detection, we aim to maximize (a) the assignment of incorrectly labeled instances to the noisy cluster while minimizing (b) the assignment of correctly labeled instances to the noisy cluster.
Intuitively, in an ideally clustered space, the difference between (a) and (b) needs to be maximized.

Since we cannot access the ground-truth annotations to compute (a) and (b), we use the most representative instances as a workaround.
Considering the corrupted dataset has a higher noise rate than the original dataset, we emulate (a) using instances predicted as noisy among the corrupted dataset, i.e., $\bar{Z}_{noisy}$.
Similarly, (b) is emulated using instances predicted as clean among the original dataset with a lower noise rate, i.e., $\widetilde{Z}_{clean}$.
Our validation metric is defined as the difference between two emulated values as
\begin{equation}
\label{eq:validmetric}
     \Big(     \sum_{\mathbf{z}_\mathbf{x}\in\bar{Z}_{noisy}} \frac{q_{noisy}(\mathbf{z_x})}{|\bar{Z}_{noisy}|} - 
     \sum_{\mathbf{z_x}\in\widetilde{Z}_{clean}} \frac{q_{noisy}(\mathbf{z_x})}{|\widetilde{Z}_{clean}|}
     \Big)^2.
\end{equation}
The larger value indicates the better clustering quality for noisy label detection.
Compared to the conventional metrics for assessing cluster separation \cite{rousseeuw1987silhouettes, davies1979cluster}, this metric is tailored for our \proposed framework and provides a more effective measure of noisy label detection efficacy.

\section{Experiments}
\label{sec:exp}

\begin{table*}[hbt!]
\centering
\fontsize{8}{8}\selectfont
\setlength{\tabcolsep}{4.0pt}
\begin{tabular}{c|ccccc|cccc|c}
\toprule
Dataset     & \multicolumn{5}{c|}{CIFAR-10}   & \multicolumn{4}{c|}{CIFAR-100}  &        \\ 
\midrule
Noise type  & Sym.    & Asym. & Inst. & Agg.  & Worst & Sym.     & Asym. & Inst. & Human & Avg.   \\
Noise rate ($\eta$) & 0.6     & 0.3   & 0.4   & 0.09    & 0.4   & 0.6      & 0.3   & 0.4   & 0.4   &        \\\midrule
Avg.Encoder & \textbf{98.0}  $\pm$ 0.03 & 89.7 $\pm$ 0.14 & 22.4 $\pm$ 33.5 & 67.3 $\pm$ 0.42 & \textbf{92.8} $\pm$ 0.11 & \textbf{96.7} $\pm$ 0.07 & 74.9 $\pm$ 0.17 & 76.8 $\pm$ 0.51 & 79.5 $\pm$ 0.31 & 77.6   \\
AUM         & 95.7 $\pm$ 0.07 & 86.5 $\pm$ 0.18 & 81.9 $\pm$ 0.72  & 74.0 $\pm$ 0.16  & 88.7 $\pm$ 0.19 & 96.4 $\pm$ 0.10 & 74.7 $\pm$ 0.21 & 81.2 $\pm$ 0.25 & 74.6 $\pm$ 1.25 & 83.7  \\
CL          & 96.6 $\pm$ 0.04 & \textbf{94.0} $\pm$ 0.10 & 82.0 $\pm$ 0.21  & 68.6 $\pm$ 0.33 &  88.3 $\pm$ 0.11 & 88.0 $\pm$ 0.08 & 68.6 $\pm$ 0.16 & 75.9 $\pm$ 0.12 & 71.9 $\pm$ 0.10 & 81.5  \\
CORES       & 97.7 $\pm$ 0.03 & 5.00  $\pm$ 0.33 & 19.2 $\pm$ 0.10  & \textbf{80.5} $\pm$ 0.09 &  77.5 $\pm$ 0.09 & 83.9 $\pm$ 0.20 & 21.9 $\pm$ 0.32 & 36.7 $\pm$ 0.41 & 36.0 $\pm$ 0.12 & 50.9  \\
SIMIFEAT-V  & 95.1 $\pm$ 0.06 & 89.4 $\pm$ 0.08 & 88.1 $\pm$ 0.11  & 79.6 $\pm$ 0.13 & 91.6 $\pm$ 0.06 & 86.0 $\pm$ 0.09 & 73.8 $\pm$ 0.07 & 80.5 $\pm$ 0.09 & 77.1 $\pm$ 0.12 & 84.6  \\
SIMIFEAT-R  & 96.1 $\pm$ 1.41 & 88.9 $\pm$ 0.14 & 91.2 $\pm$ 0.07  & 79.6 $\pm$ 0.40 &  91.7 $\pm$ 0.35 & 90.3 $\pm$ 0.07 & 68.0 $\pm$ 0.10 & 77.3 $\pm$ 0.09 & 79.3 $\pm$ 0.11 & 84.7  \\\midrule
\proposed   & \textbf{98.0} $\pm$ 0.04 & \textbf{94.0} $\pm$ 0.15 & \textbf{92.3} $\pm$ 0.38  & 79.6 $\pm$ 0.37 &  92.3 $\pm$ 0.19 & 94.3 $\pm$ 0.34 & \textbf{76.3} $\pm$ 0.23 & \textbf{81.7} $\pm$ 0.21 & \textbf{80.4} $\pm$ 0.17 & \textbf{87.7}  \\

\bottomrule

\end{tabular}
\caption{
Average F1 score (\%) along with standard deviation across ten independent runs of \proposed and baseline methods on CIFAR-10 and CIFAR-100. 
All methods except SIMIFEAT utilize the identical fixed image encoder from CLIP~\cite{radford2021learning} and train only a subsequent MLP, while SIMIFEAT uses pre-trained CLIP as a feature extractor. 
The rightmost column averages the F1 scores across nine different settings. 
``Agg.'', ``Worst'', and ``Human'' correspond to the real-world human label noises \cite{wei2021learning}.
The best results are in \textbf{bold}.
}
\label{tbl:main}
\end{table*}

\begin{table*}[hbt!]
\centering
\fontsize{8}{8}\selectfont
\setlength{\tabcolsep}{4.0pt}
\begin{tabular}{c|ccccc|cccc|c}
\toprule
Dataset     & \multicolumn{5}{c|}{CIFAR-10}   & \multicolumn{4}{c|}{CIFAR-100}  &        \\ 
\midrule
Noise type  & Sym.    & Asym. & Inst. & Agg.  & Worst & Sym.     & Asym. & Inst. & Human & Avg.   \\
\midrule

Avg.Encoder & 94.1 $\pm$ 0.14 & 85.4 $\pm$ 0.19 & 88.5 $\pm$ 0.20 & 63.6 $\pm$ 0.72 & 87.6 $\pm$ 0.18 & \textbf{92.5} $\pm$ 0.34 & 75.2 $\pm$ 0.36 & 76.0 $\pm$ 0.49 & \textbf{78.8} $\pm$ 0.18 & 82.4 \\
AUM         & 75.4 $\pm$ 0.22 & 46.4 $\pm$ 0.30 & 57.7 $\pm$ 0.03 & 16.7 $\pm$ 0.01 & 57.8 $\pm$ 0.04 & 75.8 $\pm$ 0.21 & 46.7 $\pm$ 0.32 & 57.8 $\pm$ 0.10 & 58.0 $\pm$ 0.21 & 54.7 \\
CL          & 88.7 $\pm$ 0.56 & 91.9 $\pm$ 0.12 & 82.5 $\pm$ 0.37 & 57.0 $\pm$ 0.31 & 80.0 $\pm$ 0.32 & 77.9 $\pm$ 0.39 & 62.4 $\pm$ 0.24 & 67.3 $\pm$ 0.28 & 65.2 $\pm$ 0.19 & 74.8 \\
CORES       & 92.9 $\pm$ 0.17 & 26.7 $\pm$ 0.44 & 49.2 $\pm$ 1.15 & 63.6 $\pm$ 0.58 & 74.7 $\pm$ 0.36 & 66.3 $\pm$ 0.35 & 33.8 $\pm$ 0.46 & 39.2 $\pm$ 0.45 & 31.9 $\pm$ 0.48 & 53.2 \\
SIMIFEAT-V  & \textbf{94.6} $\pm$ 0.06 & 84.7 $\pm$ 0.17 & 83.7 $\pm$ 0.08 & 69.4 $\pm$ 0.17 & 88.3 $\pm$ 0.08 & 88.0 $\pm$ 0.09 & 70.3 $\pm$ 0.14 & 77.8 $\pm$ 0.10 & 76.2 $\pm$ 0.14 & 81.4 \\
SIMIFEAT-R  & 92.9 $\pm$ 1.84 & 84.0 $\pm$ 0.13 & 86.9 $\pm$ 0.08 & 68.8 $\pm$ 0.32 & \textbf{88.5} $\pm$ 0.36 & 89.7 $\pm$ 0.07 & 66.2 $\pm$ 0.11 & 75.5 $\pm$ 0.08 & 77.8 $\pm$ 0.13 & 81.2 \\\midrule
\proposed   & 93.6 $\pm$ 0.18 & \textbf{94.2} $\pm$ 0.45 & \textbf{91.5} $\pm$ 0.31 & \textbf{72.6} $\pm$ 2.46 & 87.8 $\pm$ 0.37 & 91.3 $\pm$ 0.46 & \textbf{79.2} $\pm$ 0.59 & \textbf{79.5} $\pm$ 1.14 & 77.3 $\pm$ 0.54 & \textbf{85.2} \\

\bottomrule

\end{tabular}
\caption{
Average F1 score (\%) under identical settings to those in Table \ref{tbl:main} except for the backbone model. 
All methods except SIMIFEAT utilize a randomly initialized Renset34~\cite{he2016deep}, while SIMIFEAT uses a pre-trained ResNet34 on ImageNet~\cite{deng2009imagenet} as a feature extractor. 
}

\label{tbl:main_resnet}
\end{table*}

\subsection{Experiment setup}
\smallsection{Datasets} We evaluate the performance of \proposed on benchmark datasets with different types of label noise, originating from diverse sources: (1) synthetic noise on CIFAR-10 and CIFAR-100  \cite{krizhevsky2009learning}, (2) real-world human noise on CIFAR-10N and CIFAR-100N \cite{wei2021learning}, and (3) systematic noise\footnote{In case of Clothing1M, systematic noise is induced by automatic annotation from the keywords present in the surrounding text of each image.} on Clothing1M \cite{xiao2015learning}.
In the case of synthetic noise, following the previous experimental setup \cite{zhu2022detecting}, we artificially introduce the noise by using different strategies with specific noise rates $\eta$ as outlined below.
\begin{itemize}
    \item \textbf{Symmetric Noise} (Sym., $\eta=0.6$) randomly replaces the label with one of the other classes.
    \item \textbf{Asymmetric Noise} (Asym., $\eta=0.3$) performs pairwise label flipping, where transition can only occur from a given class $i$ to the next class $(i \ \mathrm{mode}\ C)+1$. 
    \item \textbf{Instance-dependent Noise} (Inst., $\eta=0.4$) changes labels based on the transition probability calculated using instance's corresponding features \cite{xia2020part}. 
\end{itemize}
In the case of human noise, we choose two noise subtypes for CIFAR-10N (denoted by Agg. and Worst) and a single noise subtype for CIFAR-100N (denoted by Human). 
More details of the datasets are presented in Appendix A.1.

\smallsection{Baselines} 
We compare \proposed with various noisy label detection methods. 
All the methods except SIMIFEAT use training signals to identify incorrectly labeled instances.
\begin{itemize}
     \item \textbf{\avgenc} is a naive baseline that discriminates between clean and noisy labels by using a one-dimensional Gaussian mixture model \cite{zoran2011learning} on the averaged training signals (i.e., logit difference) over the epochs. 

    \item \textbf{\aum} \cite{pleiss2020identifying} uses summation of training signals (i.e., logit difference) over the epochs and identifies correctly/incorrectly labeled instances based on a threshold. 
    
    \item \textbf{\cl} \cite{northcutt2021confident} uses a predicted probability of the given label (i.e., confidence) and filter out the instances with low confidence based on class-conditional thresholds. 
    
    \item \textbf{\cores} \cite{cheng2020learning} leverages a training loss for noisy label detection, progressively filtering out incorrectly labeled instances using its proposed sample sieve. 
    
    \item \textbf{\simifeat} \cite{zhu2022detecting} is a training-free approach that effectively detects noisy labels by utilizing $K$-nearest neighbors in the feature space of a pre-trained model.
\end{itemize} 

\smallsection{Implementation details}
For our label corruption process, we use the corruption rate $\gamma=0.1$ as the default. 
To generate the training dynamics, we employ DNN classifiers: 
ResNet34~\cite{he2016deep} and the pre-trained ViT-B/32-CLIP~\cite{radford2021learning} with a multi-layer perceptron (MLP) of two hidden layers. 
To encode the training dynamics, we use a three-layered 1D-CNN architecture~\cite{wang2017time} as the dynamics encoder.
The hyperparameter $\alpha$ is selected as either 0.05 or 0.5.
For more details about implementation, please refer to Appendix A.2.

\subsection{Noisy label detection performance}
\label{subsec:detperf}
We first evaluate \proposed and the baseline methods for noisy label detection.
Table \ref{tbl:main} and Table \ref{tbl:main_resnet} present their detection F1 scores for two classifiers, CLIP w/ MLP and ResNet34, across various noise types and rates.
Notably, \proposed achieves the best performance on average, i.e., $+$3.0\% in Table \ref{tbl:main} and $+$2.8\% in Table \ref{tbl:main_resnet}, demonstrating its robustness to various types of noisy conditions. 
On the other hand, the baseline methods relying on training signals (i.e., \avgenc, \aum, \cl, and \cores) show considerable variations in performance across different noise types.
For example, in the case of CIFAR-10, \avgenc and \cores perform well for symmetric noises, whereas they struggle with identifying asymmetric or instance noises.
It is worth noting that asymmetric and instance noise are more complex than symmetric noise in that they can have a more detrimental impact on model performance~\cite{oyen2022robustness}.
These results strongly support the superiority of our \proposed framework in handling a wide range of label noise variations.

\subsection{Effectiveness of validation metric}

\begin{table}[t]
\centering
{
\fontsize{7}{7}\selectfont
\resizebox{0.99\linewidth}{!}{
\setlength{\tabcolsep}{4.0pt}

\begin{tabular}{c|cc|cc}
    \toprule
    % \hline
      \multirow[b]{2}{*}{\shortstack[t]{Validation\\ metric}} & \multicolumn{2}{c|}{CIFAR-10} &  \multicolumn{2}{c}{CIFAR-100} \\ 
      % \cmidrule{2-7}
         & Inst. & Agg.   & Inst. & Human    \\
     \midrule
     
Max epoch  & 86.7 $\pm$ 6.75 & 77.8 $\pm$ 3.35   & 61.0 $\pm$ 10.3  & 64.3 $\pm$ 4.40 \\
DBI           & 86.3 $\pm$ 8.75 & 76.7 $\pm$ 3.91 & 60.0 $\pm$ 10.2  & 64.8 $\pm$ 9.70\\
Ours         & 92.3 $\pm$ 0.38& 79.6 $\pm$ 0.37   & 81.7 $\pm$ 0.21 & 80.4 $\pm$ 0.17\\ \midrule
Opt epoch     & 92.6 $\pm$ 0.40 & 80.40 $\pm$ 0.44   & 81.8 $\pm$ 0.08 & 80.5 $\pm$ 0.18   \\
    \bottomrule
\end{tabular}
}
}
\caption{F1 score (\%) of our dynamics encoder over various validation metrics on CIFAR-10 and CIFAR-100 using CLIP w/ MLP as a classifier.}
\label{tbl:validation_metric}
\end{table}

\begin{figure}[t]
    \centering
    
    \begin{subfigure}[b]{\linewidth}
        \centering
        \includegraphics[width=0.85\textwidth]{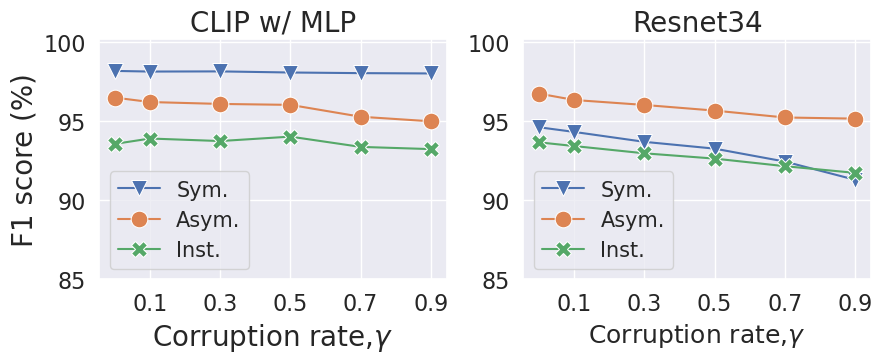}
        \caption{Supervised setting}
        \label{fig:ablation_cor_sup}
    \end{subfigure}
    \\ % Creates a line break between the two subfigures

    \begin{subfigure}[b]{\linewidth}
        \centering
        \includegraphics[width=0.85\textwidth]{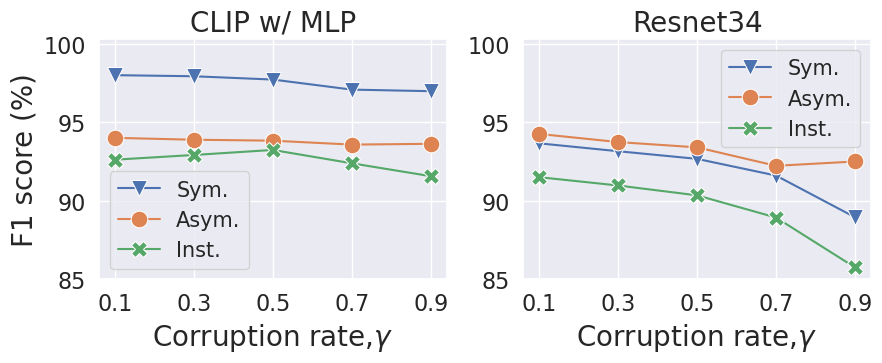}
        \caption{Unsupervised setting: \proposed.}
        \label{fig:ablation_cor_unsup}
    \end{subfigure}
    \\ % Creates a line break between the two subfigures

    \caption{F1 score (\%) changes with respect to corruption rate $(\gamma)$ on CIFAR10 in supervised and unsupervised settings using CLIP w/ MLP (Left) and ResNet34 (Right) as classifiers.   }
    \label{fig:ablation_cor}
\end{figure}

\label{subsec:effectvalid}
To demonstrate the effectiveness of the proposed validation metric (Sec.\ref{subsubsec:validating}), we compare the detection performance of our dynamics encoder by employing our proposed metric and alternative criteria as stopping conditions during the training.
\textbf{Max epoch} signifies the training over the maximum number of epochs.
\textbf{Davies-Bouldin Index (DBI)}~\cite{davies1979cluster} assesses the quality of clustering results by calculating the ratio of intra-cluster distances to inter-cluster separations. 
A lower DBI value implies more compact and well-separated clusters, i.e., better clustering quality.
In addition, \textbf{Opt epoch} selects the optimal training epoch that achieves the best detection results, providing the upper bound of detection performance.

In Table \ref{tbl:validation_metric}, 
our performance is close to the optimal case across various noise types and datasets,
whereas Max epoch and DBI fail to stop the training process at a proper epoch on CIFAR-100.
In conclusion, using the proper validation metric is critical for achieving competitive detection performance, particularly in the scenario where ground-truth annotations are not available for validation.

\subsection{Quantitative analyses}
\label{subsec:ablation}

\smallsection{The effect of corruption rate}
We analyze the effect of increasing the corruption rate, which in turn amplifies the overall noise level.\footnote{The overall noise rate is formulated as $\eta_{over}=\frac{\eta +\gamma\cdot\eta_\gamma}{1+\gamma}$.} 
For thorough analyses, we conduct a controlled experiment within a supervised framework using classification,\footnote{See Appendix B.1 for the details.} assuming the availability of ground-truth annotations that indicate each instance as being correctly or incorrectly labeled. 
We then compare these results, generally regarded as the performance upper bound for unsupervised methods, with those obtained by an unsupervised approach.
We focus on assessing the ability of our proposed unsupervised learning model, i.e., \proposed, to discriminate training dynamics and how this discrimination is affected by increasing the overall noise level through corruption.

As shown in Figure \ref{fig:ablation_cor}, the detection F1 scores achieved by \proposed (Figure \ref{fig:ablation_cor_unsup}) approaches those of supervised learning (Figure \ref{fig:ablation_cor_sup}), demonstrating the effectiveness of training dynamics. 
This proximity is especially notable when utilizing a powerful image encoder, i.e., CLIP, which makes the training dynamics less susceptible to changes in the corruption rate. 
In contrast, the training dynamics from ResNet34 are more affected by increased corruption rate.
Surprisingly, in the case of ``Inst.'' type label noise, the training dynamics from the CLIP w/ MLP classifier become even more distinguishable as the corruption rate increases to 0.5.
It shows that a higher noise rate in the training dataset can enhance the discernibility of the training dynamics.
We hypothesize that the symmetric noise introduced through our label corruption process may reduce the overall difficulty of the detection task.
This is consistent with the assertion in Sec. \ref{subsubsec:clustering} that the symmetric noise is relatively straightforward to identify and, in turn, contributes to improving the performance of noisy label detection.

\begin{table}[t]
\centering
{
\fontsize{8}{8}\selectfont
\setlength{\tabcolsep}{4.0pt}
\resizebox{0.99\linewidth}{!}{
\begin{tabular}{cc|ccc}
    \toprule
     \footnotesize $\mathcal{L}_{cluster}$ & \footnotesize $\mathcal{L}_{align}$  & Asym. & Inst.   & Agg.    \\
     \midrule
          &                          &	93.8 $\pm$ 0.17	 & 91.8 $\pm$ 0.39  &  78.8 $\pm$ 0.37   \\
    $\checkmark$  &                  &    93.2 $\pm$ 0.11    & \textbf{92.7} $\pm$ 0.36   &	76.8 $\pm$ 0.83   \\
     $\checkmark$ &  $\checkmark$    &    \textbf{94.0} $\pm$ 0.15    &	92.3 $\pm$ 0.38   &	\textbf{79.6} $\pm$ 0.37\\  
    \bottomrule
\end{tabular}
}
}
\caption{F1 score (\%) of \proposed that ablates the clustering and alignment loss on CIFAR10 using CLIP w/ MLP as a classifier.
The first row reports the detection performance with a randomly initialized dynamics encoder.}
\label{tbl:ablation_loss}
\end{table}

\smallsection{The effect of two losses} 
We examine the effect of the clustering and alignment losses within our \proposed framework.
In Table~\ref{tbl:ablation_loss}, both losses enhance detection performance. 
We also observe that the alignment loss effectively addresses the high imbalance between clean and noisy instances, particularly in scenarios with a low noise rate (e.g., ``Agg.'' on CIFAR-10). 
Given that \proposed intentionally increases the noise rate by augmenting instances with corrupted labels, its benefits become more pronounced when dealing with datasets featuring a small original noise rate. 
In such cases, the alignment loss is crucial in stabilizing the clustering process by aligning the distinct distributions of original and corrupted instances.

{\subsection{Compatibility analyses with robust learning }
\label{subsec:companal}

\begin{figure}[t]
    \centering
    
    \begin{subfigure}[b]{\linewidth}
        \centering
        \includegraphics[width=0.85\textwidth]{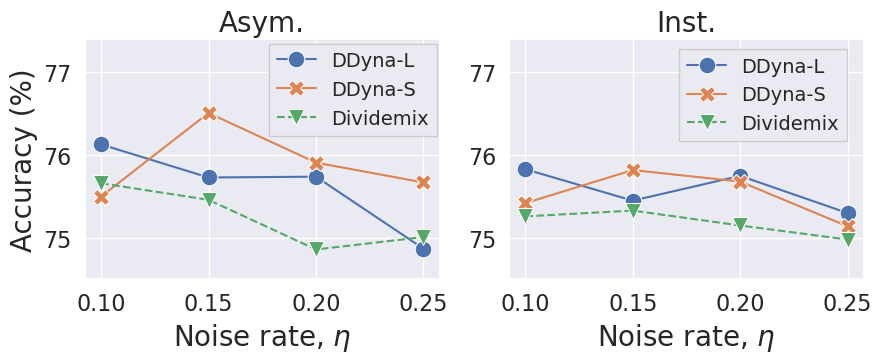}
        \caption{Classification accuracy (\%) of robust learning}
        \label{fig:divide_acc}
    \end{subfigure}
    \\ % Creates a line break between the two subfigures

    \begin{subfigure}[b]{\linewidth}
        \centering
        \includegraphics[width=0.85\textwidth]{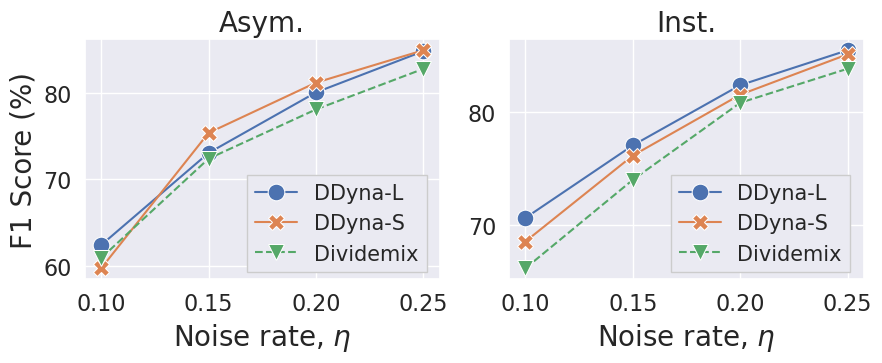}
        \caption{Noisy label detection F1 score (\%)}
        \label{fig:divide_f1}
    \end{subfigure}
    \\ % Creates a line break between the two subfigures

    \caption{Compatibility analysis of \divmix with \proposed on CIFAR100 over ``Asym.'' and ``Inst.'' with respect to noise rate}
    \label{fig:divide}
\end{figure}

We investigate the compatibility and synergistic effects of integrating our framework with various robust learning techniques: a semi-supervised approach (\divmix \cite{li2020dividemix}), loss functions (GCE~\cite{zhang2020distilling} and SCE~\cite{wang2019symmetric}), and a regularization method (ELR~\cite{liu2020early}). 
Detailed analyses of incorporating the loss functions and regularization technique on the Clothing1M dataset are provided in Appendix D.

For the semi-supervised approach, we select \divmix \cite{li2020dividemix} that iteratively detects incorrectly labeled instances and treats them as \textit{unlabeled} instances. 
We construct integrated models of \divmix and \proposed through two distinct approaches: (1) \textbf{\divl} is leveraging \divmix to obtain the training dynamics of both original and corrupted datasets within our framework, and (2) \textbf{\divs} is substituting the original detection method in \divmix, i.e., GMM, with \proposed. 
For the base architecture, we employ an 18-layer PreAct ResNet~\cite{he2016identity}, adhering to its default optimization settings and hyperparameters, as specified in the original paper \cite{li2020dividemix}.

\smallsection{Classification accuracy} 
We explore the impact of our framework on the classifier's accuracy, specifically introducing a corrupted dataset (\divl) and supplanting the existing noise detection method (\divs).
Figure \ref{fig:divide_acc} demonstrates that both enhance classification performance.
In essence, results obtained with \divl demonstrate that instances with symmetric label noise introduced through our corruption process prove beneficial for noise robust learning, especially in scenarios featuring a low noise rate in the original dataset, pointed out as a challenging setting for \divmix~\cite{wei2021learning}.

\smallsection{Detection F1 score}
To report the noisy label detection performance within robust learning framework, i.e., \divmix and \divs, we measure F1 score at every epoch and report the value when test classification accuracy is at its highest.
Note that they leverage a clean test dataset to identify the optimal detection point; on the contrary, the noisy detection method (\divl) operates without access to clean data, instead employing the procedure for model validation on the noisy dataset itself (Sec. \ref{subsubsec:validating}), presenting a more challenging task.   
Figure~\ref{fig:divide_f1} indicates that \divs and \divl further improves the detection F1 score of \divmix, indicating the great compatibility of \proposed with existing semi-supervised noise robust learning. 
In scenarios involving ``Inst.'' label noise, \divl exhibits compelling synergistic effects across a wide range of noise rates.

\section{Conclusion}
\label{sec:conclusion}

This paper proposes a new \proposed framework that distinguishes incorrectly labeled instances from correctly labeled ones via clustering of their training dynamics.
\proposed first introduces a label corruption strategy that augments the original dataset with intentionally corrupted labels, enabling indirect simulation of the model's behavior on noisy labels.
Subsequently, \proposed learns to induce two clearly distinguishable clusters for clean and noisy instances by enhancing the cluster cohesion and alignment between the original and corrupted dataset.
Furthermore, \proposed adopts a simple yet effective validation metric to indirectly estimate its detection performance in the absence of annotations of clean and noisy labels.
Our comprehensive experiments on real-world datasets demonstrate the detection efficacy of \proposed, its remarkable robustness to various noise types and noise rates, and great compatibility with existing approaches to noise robust learning.   

\section{Acknowledgements}
\label{sec:acknowledgements}

This work was supported by the IITP grant funded by the MSIT (No.2018-0-00584, 2019-0-01906, 2020-0-01361), the NRF grant funded by the MSIT (No.2020R1A2B5B03097210, RS-2023-00217286), and the Digital Innovation Hub project supervised by the Daegu Digital Innovation Promotion Agency (DIP) grant funded by the Korea government (MSIT and Daegu Metropolitan City) in 2024 (No. DBSD1-07).

{
    \small
    \bibliographystyle{ieeenat_fullname}
    \bibliography{main}
}

\clearpage
\appendix
\section*{Supplementary Material: ``Learning Discriminative Dynamics with~Label~Corruption~for~Noisy~Label~Detection''}

\section{Experiment Setup}

\subsection{Datasets} 
\smallsection{Synthetic noise: instance-dependent label noise} 
We detail the process of generating instance-dependent label noise~\cite{xia2020part}, which is the synthetic type label noise utilized in our experiments. 
The key idea is that the probability of an instance being incorrectly labeled to other classes is calculated based on both the input feature and its label, using randomly generated feature projection matrices with respect to each class. 
The procedure is provided in Algorithm \ref{alg:algorithm}. 
\begin{algorithm}
\caption{Instance-Dependent Label Noise Synthesis}
\label{alg:algorithm}
\textbf{Input}: Clean dataset \( D=\{(\mathbf{x}_n, y_n)\}^N_{n=1} \), \( \mathbf{x}_n\in\mathbb{R}^{d_{\mathbf{x}}} \), Noise rate \( \eta \), Number of classes \( C \) \\
\textbf{Output}: Noisily labeled dataset \( \tilde{D}=\{(\mathbf{x}_n, \tilde{y}_n)\}^N_{n=1} \)
\begin{algorithmic}[1]
    \STATE Sample \( C \) feature projection matrices \{\( \mathbf{W}_1 \), ...,\( \mathbf{W}_C \)\} from a standard normal distribution \( \mathcal{N}(0,1) \), with each \( \mathbf{W}_c \in \mathbb{R}^{d_{\mathbf{x}} \times C} \).
    \FOR{\( n=1,\ldots,N \)}
        \STATE Sample \( q \in \mathbb{R}\) from a truncated normal distribution \( \mathcal{N}(\eta, 0.1^2) \) within the interval [0,1].
        \STATE Compute probability vector by \( p = \mathbf{x}_n \mathbf{W}_{y_n} \in \mathbb{R}^{C} \).
        \STATE Set the probability of the true class to be negative infinity \( p_{y_n} = -\infty \).
        \STATE Adjust \( p = q \times \mathrm{Softmax}(p) \) and set \( p_{y_n} = 1-q \).
        \STATE Sample corrupted label \( \tilde{y}_n \) from \( C \) classes according to the modified probability distribution \( p \).
    \ENDFOR
\end{algorithmic}
\end{algorithm}

\smallsection{Clothing1M~\cite{xiao2015learning}}
To assess \proposed's performance with systematic type label noise, we use a real-world dataset Clothing1M, which consists of clothing images across 14 classes\footnote{T-shirt, Shirt, Knitwear, Chiffon, Sweater, Hoodie, Windbreaker, Jacket, Down Coat, Suit, Shawl, Dress, Vest, and Underwear} collected from online shopping websites.
It comprises one million images with inherent noisy labels induced by automated annotations derived from keywords in the text surrounding each image.  
It also provides 50K, 14K, and 10K instances verified as clean for training, validation, and testing purposes. 
Adhering to the previous experimental setup~\cite{kim2021fine}, for training, we utilize randomly sampled 120K instances from the 1M noisy dataset while ensuring each class is balanced. 
To evaluate classification performance, we use the 10K clean test set.

% \section{Experimental Settings}
\subsection{Reproducibility}
For reproducibility, we provide detailed hyperparameters for (1) classifiers used to generate training dynamics or to learn robust models and (2) dynamics encoder to learn discriminative representations of the training dynamics.

\smallsection{Classifier}
Table \ref{tbl:parameter} shows details of the datasets, models, and training parameters used to generate training dynamics or to learn robust models in each section of this paper.
Optimizer and momentum are fixed as SGD and 0.9, respectively. 
In the case of CLIP with MLP, we obtain input features using a fixed image encoder from CLIP and train only MLP, which consists of two fully connected layers of 512 units with ReLUs~\cite{krizhevsky2012imagenet}. 
Resnet50 is pre-trained on ImageNet~\cite{deng2009imagenet} and is fine-tuned on Clothing1M.
We follow the experimental setups described in the reference papers.

\begin{table}[h]
\centering
{
\resizebox{0.99\linewidth}{!}{
\setlength{\tabcolsep}{4.0pt}

\begin{tabular}{lcccc}
    \toprule
    Dataset & \multicolumn{3}{c}{CIFAR-10/CIFAR-100} & Clothing1M\\ \midrule
    Section & \multicolumn{2}{c}{5.2 to 5.4} &  5.5   & Appendix~\ref{sec:clothing} \\ \midrule
        \raisebox{1.2ex}[0pt][0pt]{Model}
        & {\shortstack[h]{CLIP \cite{radford2021learning}\\w/ MLP}}
        & {\shortstack[h]{Resnet34\\ \cite{he2016deep, wei2021learning}}}       
        & {\shortstack[h]{PreAct-\\ Resnet18 \cite{he2016identity, li2020dividemix}}} 
        & {\shortstack[h]{Resnet50 \\  \cite{he2016deep, kim2021fine} }}                    \\ \midrule
    Learning rate   & 0.1  & 0.1   & 0.02 & 0.002 \\ 
    Weight decay & $5\times 10^{-4}$  &  $5\times 10^{-4}$  & $5\times 10^{-4}$ & 0.001 \\
    LR scheduler & Cosine  &  Multi-step  & Multi-step &  Multi-step \\
    Batch size & 128  &  128  & 128 &  64 \\
  Epochs & 30                   & 100               & 300                     & 10     \\ 
  $\alpha$ & 0.5                   & 0.05               & 0.05                     & 0.5     \\ 
   
   \bottomrule
\end{tabular}
}
}
\caption{Detailed hyperparameters used in the experiments for the classifiers.}
\label{tbl:parameter}
\end{table}

\smallsection{{Dynamics encoder}}
For the dynamics encoder in \proposed, we use a 1D Convolutional Neural Network (1D-CNN).
It consists of three convolutional layers, each incorporating rectified linear units (ReLUs)~\cite{krizhevsky2012imagenet}, followed by a linear layer with 512 output units.
For optimization, we use Adam~\cite{kingma2014adam} with a learning rate $1\times10^{-5}$ and a weight decay $5\times 10^{-4}$ without implementing a learning rate scheduler.
The model is trained for 10 epochs with a batch size of 1024.

\section{Analyses of Training Dynamics}
To assess the distinguishability of the inherent patterns manifested in the training dynamics, we conduct a controlled experiment using classification within a supervised learning framework.
This is predicated on the assumption that ground-truth annotations are available, explicitly specifying each instance as being correctly or incorrectly labeled.

We first provide preliminaries for analyses (Sec.~\ref{sec:preliminaries}). 
Then, we demonstrate the efficacy of capturing temporal patterns in training dynamics versus summarizing these dynamics into a single scalar value (Sec.~\ref{sec:temporal}) on various training signals. 
Lastly, we evaluate which training signals exhibit more distinctive patterns (Sec.~\ref{sec:various_signal}).

\subsection{Preliminaries} 
\label{sec:preliminaries}
\smallsection{Training signals}
Table \ref{tbl:train_signal} summarizes various training signals introduced in the literature. 
Given an instance $(\mathbf{x}, y)$ and a classifier $f$, let $f(\mathbf{x}) \in\mathbb{R}^{C}$ and $f_y(\mathbf{x})$ denote the output logits of an instance $\mathbf{x}$ for $C$ classes and its value for class $y$, respectively. 
$\ell(\cdot,\cdot)$ is a loss function, and $p_{y}(\mathbf{x})= \frac{\exp{f_{{y}}(\mathbf{x})}}{\sum_{c=1}^{C}\exp{f_c(\mathbf{x})}}$ is a predicted probability of class $y$.
$\mathbf{v}_{\mathbf{x}}$ indicates penultimate layer representation vectors of an instance $\mathbf{x}$, and $\mathbf{u}_y$ is a representative vector for class $y$, derived through performing eigen decomposition on the gram matrix of data representations. 
$\langle \cdot, \cdot\rangle$ denotes inner product. 

\begin{table}[htbp]
\centering
{
\setlength{\tabcolsep}{8.0pt}
\resizebox{0.99\linewidth}{!}{
% \begin{tabular}{ll}
\begin{tabular}{p{4.5cm}p{4.5cm}}
\toprule
 Training signal & Formula, $t_{\mathbf{x}}$ \\ \midrule
 Loss~\cite{jiang2018mentornet}& $\ell(f(\mathbf{x}), y)$ \\
 Probability~\cite{chen2023sample}& $p_{y}(\mathbf{x})$ \\
 Probability difference~\cite{torkzadehmahani2022label}&$\max_{c}p_c(\mathbf{x})-p_{y}(\mathbf{x})$  \\
 Logit difference~\cite{pleiss2020identifying} & $f_{y}(\mathbf{x}) - \max_{c\neq y}f_c(\mathbf{x})$  \\
 Alignment of pre-logits~\cite{kim2021fine}& $\langle\mathbf{u}_y,\; \mathbf{v}_\mathbf{x}\rangle^2$ \\ \bottomrule
\end{tabular}
}
}
\caption{Various types of training signals.}
\label{tbl:train_signal}
\end{table}

\begin{figure}[htbp]
    \centering
    \includegraphics[scale=0.65]{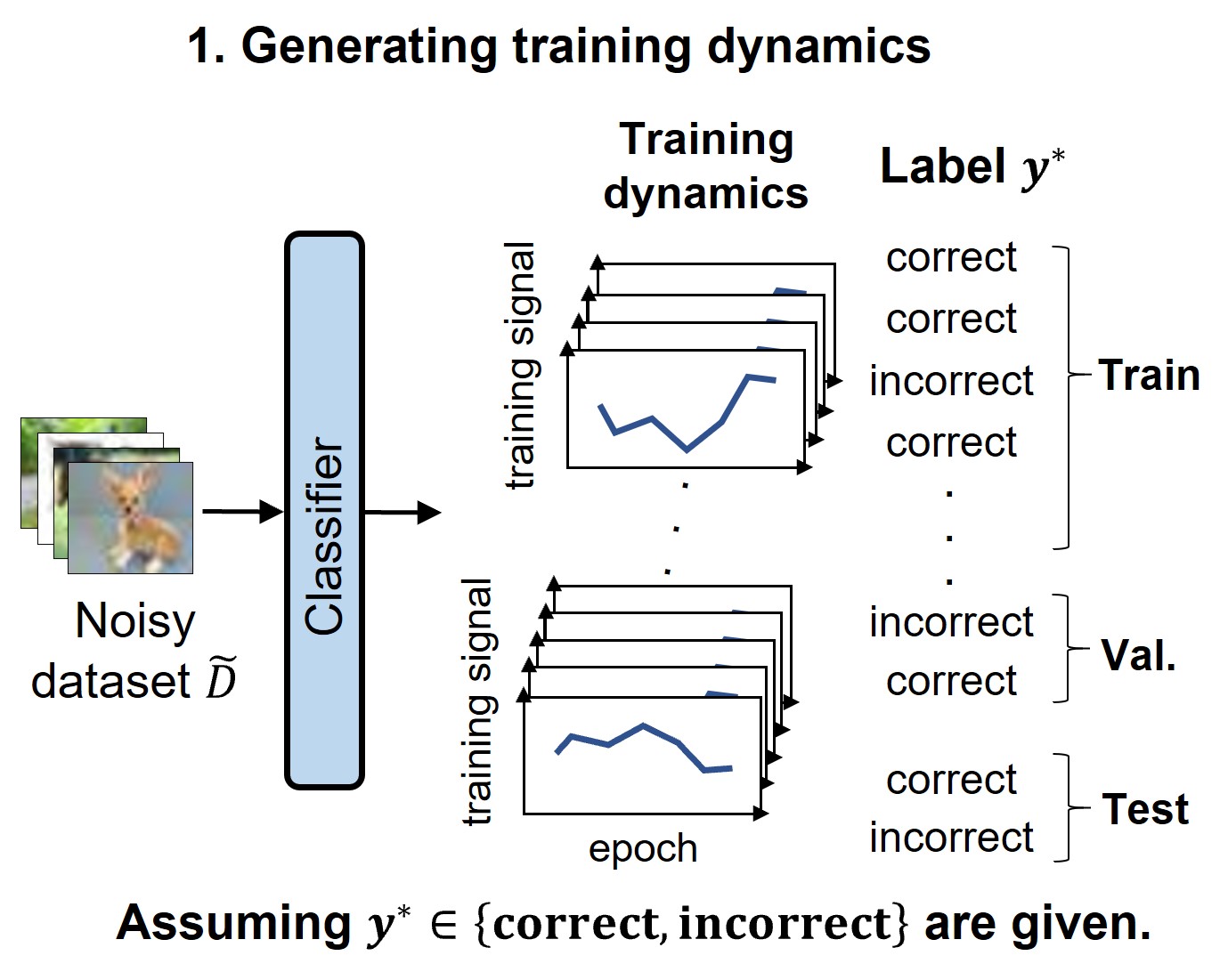}
    \caption{
    Dataset construction for supervised learning.}
    \label{fig:supervised}
\end{figure}

\smallsection{Supervised experimental setting}
As illustrated in Figure \ref{fig:supervised}, we generate training dynamics by employing a classifier that predicts the class probabilities for each input instance across the set of classes. 
Subsequently, we construct a new dataset comprising these extracted training dynamics and the corresponding ground-truth labels that are assumed to exist.
This new dataset is then utilized to train a 1D convolutional neural network (1D-CNN) classifier (henceforth referred to as a \textit{binary classifier}) that distinguishes between correctly and incorrectly labeled instances based on the patterns in their training dynamics.
We train the \textit{binary classifier} (whose encoder is the same as our dynamics encoder) for 20 epochs using the Adadelta~\cite{zeiler2012adadelta} optimizer with an initial learning rate of 1 and a StepLR scheduler that reduces it by 1\% for every epoch. 
The batch size is set to 128. 
During training, we monitor the model's performance on a validation set and report the F1 score for detecting incorrectly labeled instances on the test set, corresponding to the point where the validation F1 score achieves its maximum value.

\subsection{Temporal patterns in training dynamics} 
\label{sec:temporal}
To assess the effectiveness of capturing temporal patterns within training dynamics compared to summarizing them into a single scalar value~\cite{pleiss2020identifying, chen2023sample}, we conduct experiments using them as input to the \textit{binary classifier} in the supervised setting. 
For the training dynamics, we use 

\begin{equation}
\label{eq:dynamics_trajectories1}
     \mathbf{t}_{\mathbf{x}}=[t^{(1)}_{\mathbf{x}},..,t^{(E)}_{\mathbf{x}}], 
\end{equation}
where $t_{\mathbf{x}}^{(e)}$ is a training signal at epoch $e$ for an instance $\mathbf{x}$, and $E$ is the maximum number of training epochs.
For the summarized one, we use a statistical method~\cite{pleiss2020identifying, chen2023sample} that average the series of temporal signals into a single scalar value $s_{\mathbf{x}}$ to encapsulate the essential features.
\begin{equation}
s_{\mathbf{x}} = \frac{1}{E}\sum_{e=1}^{E}t_{\mathbf{x}}^{(e)},
\end{equation}
To evaluate the relative efficacy of these approaches, we use two distinct types of training signals: probability and logit difference in Table \ref{tbl:train_signal}.  
For the \textit{binary classifier} of the summarized one, we adopt a multi-layer perceptron (MLP) of two hidden layers.
To ensure the model's sufficient capacity to learn patterns in the data, we increase the model parameters until performance does not improve further.

\begin{figure}[htbp]
    \centering
    \includegraphics[scale=0.37]{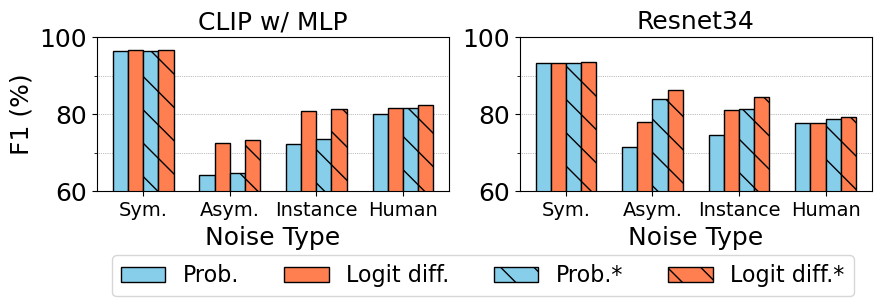}
    \caption{
Comparison of detection F1 score (\%) achieved by the \textit{binary classifiers} trained using the training dynamics (comb-pattern bar and star marker in legend) versus those trained with the summarized one for various noise types on CIFAR-100. 
Prob. and Logit diff. indicate the types of training signals in Table \ref{tbl:train_signal}.    
Noise rates of Sym., Asym., and Instance are 0.6, 0.4, and 0.3, respectively. 
The human-induced noise has noise rates of 0.4.
CLIP w/ MLP (Left) and Resnet34 (Right) are used for training dynamics generation.
}
    \label{fig:dynamics}
\end{figure}

Figure \ref{fig:dynamics} shows that the models trained with the training dynamics consistently outperform those with the summarized training dynamics. 
The results demonstrate that temporal patterns within training dynamics help distinguish between correctly and incorrectly labeled instances. 

\subsection{Comparison of various training signals} 
\label{sec:various_signal}
We compare the detection F1 score of the \textit{binary classifier} trained with the training dynamics derived from various training signals in the supervised setting. 

\begin{figure}[htbp]
    \centering
    \includegraphics[scale=0.45]{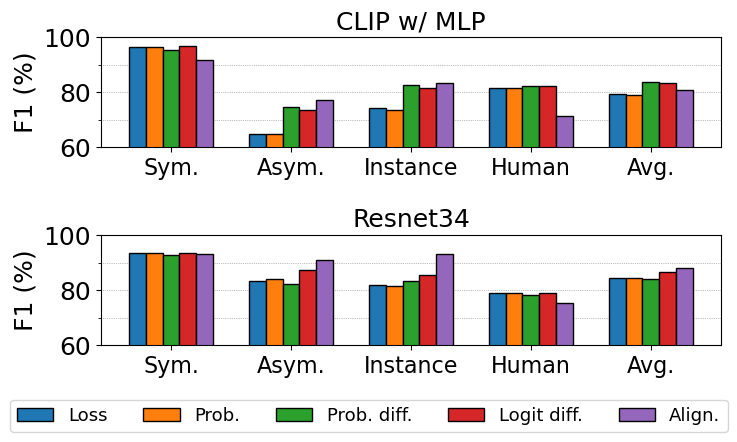}
    \caption{
Comparison of detection F1 score (\%) of the raw training dynamics from various training signals on CIFAR-100.     
Noise rates of Sym., Asym., and Instance are 0.6, 0.4, and 0.3, respectively. 
The human-induced noise type has noise rates of 0.4.
The Avg. indicates an averaged F1 score (\%) over all noise types. 
CLIP w/ MLP (Upper) and Resnet34 (Lower) are used for training dynamics generation.
 }
    \label{fig:training_signal}
\end{figure}
Figure \ref{fig:training_signal} shows that, on average, more processed training signals, such as probability differences and alignment of pre-logits, exhibit superior performance compared to simpler ones.
In this study, we select logit difference as the base proxy measure due to its consistent performance across various experimental settings. 
Moreover, we observe that detection performance for different types of noises is highly correlated with model architecture. 
We leave the study of the influence of model architectures in future work.

\section{Proof of the Lower Bound of $\eta_\gamma$}

\begin{proposition}[Lower bound of $\eta_\gamma$]
\label{pro2}
Let $\eta_{\gamma}$ denote the noise rate of the corrupted dataset. 
Given the diagonally dominant condition, i,e., $\eta < 1-\frac{1}{C}$,  for any $\gamma\in{\left(0,1\right]}$, $\eta_{\gamma}$ has a lower bound of $1-\frac{1}{C}$.
\end{proposition}

\noindent\textit{Proof.} The proportion of the correctly labeled instances in the corrupted dataset can be derived by multiplying the noise rate $\eta$ of the original dataset by the probability that a noisy label is subsequently restored to its clean label due to the corrupting process, i.e., $\eta(\frac{1}{C-1})$.
This derivation holds because the corruption process randomly flips class labels to one of the other classes uniformly.
Consequently, the noise rate $\eta_\gamma$ of the corrupted dataset is calculated as 
\begin{equation}\label{eq: eq_eta_gamma}
    \eta_\gamma = 1 - \eta\left(\frac{1}{C-1}\right).   
\end{equation}
Then, by the diagonally dominant condition, i.e., $\eta < 1-\frac{1}{C}$, Eq.~\eqref{eq: eq_eta_gamma} implies
\begin{equation}\label{eq:lower_bound}
    1-\frac{1}{C} < \eta_\gamma
\end{equation}
With this, we can derive that the corrupted dataset has a higher noise rate than the original dataset, i.e., $\eta < \eta_{\gamma}$.
Besides, we present the formulation of the overall noise rate of the original and corrupted datasets as
\begin{equation}\label{eq:overall_noise}
    \eta_{over}=\frac{\eta +\gamma\cdot\eta_\gamma}{1+\gamma}.   
\end{equation}

\section{Compatibility analysis with robust learning on Clothing 1M dataset}
\label{sec:clothing}
We also investigate the compatibility of \proposed with various loss functions (GCE~\cite{zhang2018generalized}, and SCE~\cite{wang2019symmetric}) and regularization technique (ELR~\cite{liu2020early}), specifically designed for noise robust learning.
To this end, we measure the test accuracy of such noise robust classifiers trained using the original Clothing1M dataset and the cleansed dataset (i.e., the one with only correctly labeled instances identified by \proposed), respectively.

\begin{table}[htbp]
\centering
{
\fontsize{7}{7}\selectfont
\resizebox{0.8\linewidth}{!}{
\begin{tabular}{c|ccc}
    \toprule
        {Loss type} &  GCE \cite{zhang2018generalized} & SCE \cite{wang2019symmetric}  & ELR \cite{liu2020early}  \\\cmidrule{1-4}
        Original &  71.82 & 71.75 & 72.57 \\
        Cleansed &  \textbf{72.23} & \textbf{72.37} & \textbf{73.06} \\
    \bottomrule
\end{tabular}
}
}
\caption{Classification accuracy (\%) on Clothing1M, trained with noise robust loss functions (GCE, SCE) and regularization technique (ELR) by using the original and cleansed sets, respectively.}
\label{tbl:cleansing}
\end{table}

In Table~\ref{tbl:cleansing}, we can observe consistent improvement in classification performance by cleansing the original dataset based on the detection results from \proposed, even in case the classifier is trained with a noise-robust loss function or regularization technique.

\end{document}